\documentclass{article}

\usepackage[preprint]{neurips_2026}

\usepackage[utf8]{inputenc} 
\usepackage[T1]{fontenc}    
\usepackage{hyperref}       
\usepackage{url}            
\usepackage{booktabs}       
\usepackage{amsfonts}       
\usepackage{nicefrac}       
\usepackage{microtype}      
\usepackage{xcolor}         
\usepackage{amsmath}        
\usepackage{amssymb}        
\usepackage{graphicx}       
\usepackage{colortbl}       
\usepackage{multirow}       
\usepackage[normalem]{ulem} 

\newif\ifrevfinal
\revfinaltrue
\ifrevfinal
  \newcommand{\rev}[1]{#1}
  \newcommand{\new}[1]{#1}
  \newcommand{\del}[1]{\ignorespaces}
\else
  \newcommand{\rev}[1]{\textcolor{blue}{#1}}
  \newcommand{\new}[1]{\textcolor{green!55!black}{#1}}
  \newcommand{\del}[1]{\textcolor{red}{\sout{#1}}}
\fi
\definecolor{updgreen}{HTML}{1E8449}
\ifrevfinal\newcommand{\upd}[1]{#1}\else\newcommand{\upd}[1]{\textcolor{updgreen}{#1}}\fi

\newcounter{algorithm}

\newcounter{algline}[algorithm]
\newlength{\algindentunit}

\definecolor{dpgreen}{HTML}{27AE60}   
\definecolor{mpgreen}{HTML}{82E0AA}   
\definecolor{lpgreen}{HTML}{D5F5E3}   
\definecolor{lpred}{HTML}{FADBD8}     
\definecolor{mpred}{HTML}{F1948A}     
\definecolor{dpred}{HTML}{E74C3C}     

\title{GAAMA: Graph Augmented Associative Memory for Agents}

\author{
  Swarna Kamal Paul \\
  Nagarro \\
  \texttt{swarna.paul@nagarro.com} \\
  \And
  Shubhendu Sharma \\
  Nagarro \\
  \texttt{shubhendu.sharma@nagarro.com} \\
  \And
  Nitin Sareen \\
  Nagarro \\
  \texttt{nitin.sareen@nagarro.com} \\
}

\begin{document}
\raggedbottom 

\maketitle

\begin{abstract}
\rev{AI agents that talk with users across many sessions need long-term memory to stay coherent and personalized.
Existing approaches use either flat retrieval-augmented generation (RAG) that ignores how memories connect, or entity-centric knowledge graphs.
\new{Entity-centric graphs split each message into separate triples, which loses its context, and in conversations the speakers become hubs linked to almost every message, so propagation through them tells retrieval little.}
We present GAAMA, a graph-augmented associative memory that organizes conversations around topics. It keeps verbatim episodes, uses an LLM to distill them into atomic facts and topic concepts, and then synthesizes higher-order reflections.
These four node types are linked by five kinds of structural edges, and topic concepts supply the cross-cutting paths, so mega-hubs do not form.
Retrieval blends nearest-neighbor similarity search with Personalized PageRank whose transitions depend on edge type.
We also introduce GRAFT (Graph Repair by Augmenting Facts \& Topology), an offline repair step: for questions whose retrieved evidence fell short, a bounded ReAct agent searches memory again and records the grounded answer as a fresh fact, so later questions find it directly without extra query-time cost.
On LoCoMo-10, GAAMA achieves the highest average score of all compared methods (79.1\%) and is competitive in every question category; GRAFT adds a small further gain.
\new{In a controlled comparison, replacing GAAMA's concept-linked facts with entity-relation triples lowers reward by 3.0~pp.}
\upd{On MemoryArena, scored on what memory alone determines, GAAMA has the best score on average.}}
\end{abstract}

\section{Introduction}
\label{sec:introduction}

\new{AI agents that work with users across many sessions need long-term memory to recall earlier issues, preferences, and events.
Feeding the whole history into the prompt is costly and does not scale, and flat retrieval-augmented generation (RAG)~\citep{lewis2020rag} retrieves text chunks by similarity alone, ignoring how memories connect.}

Graph-based approaches like HippoRAG~\citep{gutierrez2024hipporag} construct entity-centric knowledge graphs from OpenIE triples and retrieve via Personalized PageRank, but their entity-centric \new{approach} causes context loss during indexing and inference\new{, a limitation} that HippoRAG~2~\citep{gutierrez2025hipporag2} itself \new{calls} ``a critical flaw.''
\rev{\new{In multi-session conversations this flaw grows.
Splitting each message into triples separates an event from its circumstances, such as when it happened.
And because almost every turn involves one of the speakers, each speaker node ends up linked to most of the conversation, so propagation through it gives every turn about the same boost and does little to rank the right one.}}

\new{We present GAAMA with three contributions.}

\begin{enumerate}
\item \textbf{Hierarchical multi-level \rev{memory}.}
\new{GAAMA encodes each conversation as verbatim \emph{episodes}, self-contained \emph{facts} distilled by an LLM, and \emph{reflections} that synthesize patterns across facts, so each question can draw on the level that suits it.}
\rev{Even without graph-based retrieval, this memory outperforms flat RAG by 2.9~pp.}

\item \textbf{\new{Linked concepts, facts, episodes and reflections.}}
\new{GAAMA stores each event as a self-contained fact and links facts through topic concepts, whereas entity graphs split messages into triples that lose context and turn the speakers into hubs. Episodes are chained, and are also connected to facts.
Retrieval re-ranks nearest-neighbor results with edge-type Personalized PageRank and, unlike HippoRAG~2, makes no LLM call.
In a controlled comparison, replacing these facts with entity-relation triples lowers reward by 3.0~pp.}

\item \textbf{\rev{GRAFT for offline graph repair.}}
\rev{GRAFT, short for Graph Repair by Augmenting Facts \& Topology, repairs the graph between sessions rather than on the answer path.}
\new{When retrieved memory falls short for a question, a bounded ReAct agent searches memory again and, if memory entails its answer, stores that answer as a lower-confidence fact linked to its sources, so later questions retrieve it at no extra query-time cost.}
\end{enumerate}

\rev{Overall, GAAMA reaches 79.1\% mean reward on LoCoMo-10, the best average among all compared methods\upd{, and on MemoryArena it has the best score on both datasets}.}

\rev{The code is available at \url{https://github.com/swarna-kpaul/gaama}.}
 
\section{Related Work}
\label{sec:related}

\subsection{Memory systems for conversational agents}

\new{HippoRAG~\citep{gutierrez2024hipporag} builds a knowledge graph of phrase nodes from OpenIE triples and runs Personalized PageRank from the nodes that match entities named in the query.
HippoRAG~2~\citep{gutierrez2025hipporag2} seeds PPR from passage nodes and from query-matched triples that an LLM filters on every query.
Both versions therefore call an LLM at query time, to extract entities or to filter triples, whereas GAAMA's retrieval uses none.
In our multi-session setting, recurring person entities in such graphs accumulate hundreds of edges and become hubs.
Zep~\citep{rasmussen2025zep} builds a temporal knowledge graph whose entity nodes connect raw episodes and community summaries, and MAGMA~\citep{jiang2026magma} represents memories in semantic, temporal, causal, and entity views and retrieves them by beam search weighted by the query's intent.
Both keep entities as nodes, whereas GAAMA links memories through topic concepts and propagates relevance with edge-type-aware PPR.
Synapse~\citep{jiang2026synapse} is the closest design.
It links conversation turns to LLM-extracted concepts and ranks them by spreading activation, but it stores only raw turns, while GAAMA adds distilled facts and reflections that keep links to their sources.
A-Mem~\citep{xu2025amem} links LLM-annotated Zettelkasten notes and evolves them as new memories arrive, but retrieves by embedding similarity without relevance propagation.
Mem0~\citep{chhikara2025mem0} extracts salient facts and updates its store through LLM operations, and its graph variant Mem0$^g$ stores entity-relation triples, which exposes it to the same entity hubs.
SimpleMem~\citep{zhang2024simplemem} compresses dialogue into self-contained memory units and retrieves them through LLM-planned semantic, lexical, and metadata queries, and Nemori~\citep{wang2025nemori} segments conversations into episodes and distills semantic memory that it retrieves by embedding similarity.
Neither uses a graph.
AgeMem~\citep{yu2025agemem} learns what to store, retrieve, update, and summarize through reinforcement learning, which is complementary to GAAMA's focus on graph-structured retrieval.
Layered memories that pair raw observations with higher-level summaries or reflections appear in Generative Agents~\citep{park2023generative}, Zep, and HyperMem~\citep{yue2026hypermem}, which groups the episodes of each topic and the facts of each episode with hyperedges, and mixed abstraction levels also help LLM-based planning agents~\citep{paul2024neoplanner}.
Generative Agents scores each memory on its own and HyperMem retrieves from topics down to facts, whereas GAAMA places all levels in one graph so that relevance flows between episodes, facts, and reflections during retrieval.
Think-in-Memory~\citep{liu2023tim} stores thoughts the model generates after answering, and MemMA~\citep{lin2026memma} repairs memory with facts derived from probe questions it generates itself, whereas GRAFT stores an answer only after memory entails it, together with its provenance.}

\subsection{\new{Retrieval-augmented generation and graph retrieval}}

\new{RAG~\citep{lewis2020rag} retrieves the passages most similar to a query, which works for single-hop questions but ignores the relations between entities and events.
GraphRAG~\citep{edge2024graphrag, nan2024graphrag} builds an entity graph with community summaries over a fixed text corpus, and knowledge-graph QA systems such as QA-GNN~\citep{yasunaga2021qagnn} reason over a fixed knowledge graph such as ConceptNet.
Both add graph structure~\citep{peng2025graphragsurvey} but work on static data rather than an agent's evolving memory.
Personalized PageRank~\citep{jeh2003ppr} propagates relevance from seed nodes through a graph.
GAAMA extends it with edge-type-aware transition weights and blends PPR with semantic similarity, so the graph augments embedding retrieval rather than replacing it.}

\section{Method}
\label{sec:method}

GAAMA operates in \rev{three} phases: (A)~knowledge graph construction from conversation sessions via a three-step pipeline,  (B)~hybrid retrieval combining PPR with semantic similarity\rev{, and (C)~an offline repair pass that consolidates what retrieval could not serve}.

\subsection{Three-step knowledge graph construction}
\label{sec:kg_construction}

Given a new conversation session, GAAMA constructs a typed knowledge graph through three sequential steps that separate cheap structural operations from targeted LLM calls.

\paragraph{Step 1: Episode preservation.}
Each conversation turn is stored verbatim as an episode node, preserving the exact words from the original message without summarization or modification.
This design choice is critical to ensure no loss of information in the captured memory.
Each conversation message maps to exactly one episode node.
\rev{Consecutive episodes of a conversation are chained, including across session boundaries,} via NEXT edges encoding temporal order, \new{enabling traversal along temporal paths during retrieval}.
This step requires no LLM calls and runs in linear time.

\paragraph{Step 2: Fact and concept extraction.}
An LLM processes the episode sequence to extract two types of derived nodes:

\begin{itemize}
\item \textbf{Facts}: \rev{Self-contained atomic statements about people, events, or preferences that can be read without the original conversation}.
Facts are linked to their source episodes via DERIVED\_FROM edges, preserving provenance.
\rev{The extraction LLM also assigns each fact a belief score in $[0,1]$, close to 1 when the episodes state the fact explicitly and lower, around 0.8, when it is only implied or inferred.}

\item \textbf{Concepts}: Topic-level labels that capture the thematic content of episodes and facts.

Concepts are specifically \emph{not} person names, dates, or generic terms---they represent activities, interests, events, and topics that provide meaningful cross-cutting structure.
Episodes connect to concepts via HAS\_CONCEPT edges; facts via ABOUT\_CONCEPT edges.
\end{itemize}
\rev{Table~\ref{tab:schema} lists the four node types and five structural edge types of the resulting graph.
The schema deliberately has no entity or person node type, so names and dates appear only inside the text of episodes and facts, never as nodes.}

\paragraph{Step 3: Reflection synthesis.}
A second LLM pass synthesizes higher-order insights from multiple facts, following the concept of reflection introduced for generative agents~\citep{park2023generative}.
Reflections capture generalized patterns, preferences, or lessons that span multiple conversations.
Each reflection is linked to its supporting facts via DERIVED\_FROM\_FACT edges \rev{and carries a belief score that is higher when several consistent facts support it}.
Reflections are particularly valuable for inference-type questions that require combining information across sessions.

\begin{table}[t]
\caption{Graph schema: four node types connected by five structural edge types\rev{, with the base weight each edge type uses for PPR transitions}.}
\label{tab:schema}
\centering\small
\setlength{\tabcolsep}{4pt}
\begin{tabular}{@{}lllc@{}}
\toprule
Edge Type & Source $\to$ Target & Semantics & \rev{Weight} \\
\midrule
NEXT & Episode $\to$ Episode & Temporal succession \rev{of turns} & \rev{0.8} \\
DERIVED\_FROM & Fact $\to$ Episode & Provenance: fact extracted from episode & \rev{0.8} \\
DERIVED\_FROM\_FACT & Reflection $\to$ Fact & Provenance: reflection from facts & \rev{0.5} \\
HAS\_CONCEPT & Episode $\to$ Concept & Episode discusses topic & \rev{0.8} \\
ABOUT\_CONCEPT & Fact $\to$ Concept & Fact relates to topic & \rev{0.8} \\
\bottomrule
\end{tabular}
\end{table}

\subsection{Hybrid retrieval with Personalized PageRank}
\label{sec:retrieval}

Given a query $q$, retrieval proceeds in five steps.
\rev{Retrieval makes no LLM call. It uses only the query embedding, nearest-neighbor search, and PPR, and its seeds come directly from embedding similarity.}

\paragraph{Step 1: KNN candidate retrieval and seed selection.}
A pool of $2B$ nodes is retrieved by cosine similarity to the query embedding, where $B = \texttt{max\_facts} + \texttt{max\_reflections} + \texttt{max\_episodes}$ is the total retrieval budget.
The top-$k$ ($k{=}40$) become PPR seeds with similarity weights:
\begin{equation}
w_{\text{seed}}(n) = \text{sim}(n, q)
\label{eq:seed}
\end{equation}
Seed weights are normalized to form a probability distribution over the teleport vector.
The remaining $2B - k$ KNN candidates are retained in the candidate pool for final scoring.

\paragraph{Step 2: Graph expansion.}
Starting from seeds, edges are traversed to depth $d{=}2$, collecting all edges in the local subgraph.
Graph-discovered nodes not already in the KNN pool are fetched and added.
This surfaces structurally connected nodes that KNN may miss---e.g., a fact reachable through a shared concept via HAS\_CONCEPT $\to$ Concept $\leftarrow$ HAS\_CONCEPT.

\paragraph{Step 3: Edge-type-aware transition weights.}
Each edge from node $i$ to node $j$ with type $t$ has an effective weight $\tilde{w}_{ij} = w_{\text{base}}(t)$, where $w_{\text{base}}(t)$ is a per-type base weight (Table~\ref{tab:schema}).
Transition probabilities are computed by per-source normalization:
\begin{equation}
P_{ij} = \frac{\tilde{w}_{ij}}{\sum_{k} \tilde{w}_{ik}}
\label{eq:transition}
\end{equation}

\paragraph{Step 4: Personalized PageRank.}
Iterative PPR runs on the local subgraph with teleport vector $\mathbf{v}$ derived from seed weights and damping factor $\alpha = 0.6$:
\begin{equation}
r_j^{(t+1)} = (1 - \alpha) \cdot v_j + \alpha \sum_{i} r_i^{(t)} \cdot P_{ij} + \alpha \cdot S^{(t)} \cdot v_j
\label{eq:ppr}
\end{equation}
where $S^{(t)} = \sum_{i:\, \text{deg}(i)=0} r_i^{(t)}$ is the sink mass redistributed according to the teleport vector.
Iteration continues until convergence ($\|\mathbf{r}^{(t+1)} - \mathbf{r}^{(t)}\|_1 < 10^{-6}$) or a maximum of \rev{100} iterations.
PPR scores are max-normalized to $[0, 1]$.

\paragraph{Step 5: Additive scoring.}
The final relevance score is computed over all candidate nodes---the full $2B$ KNN pool plus graph-discovered nodes:
\begin{equation}
\text{score}(n) = b(n) \cdot \bigl(w_{\text{ppr}} \cdot \text{ppr}(n) + w_{\text{sim}} \cdot \text{sim}(n, q)\bigr)
\label{eq:score}
\end{equation}
with $w_{\text{ppr}}{=}0.1$, $w_{\text{sim}}{=}1.0$, and $b(n) \in [0, 1]$ the \emph{belief} of node $n$.
\rev{Similarity scores, like PPR scores, are max-normalized over the candidate pool before the two are combined.}

KNN candidates without PPR scores receive $\text{ppr}(n) = 0$; graph-discovered nodes without KNN similarity have their similarity computed on demand.
The low PPR weight ensures graph structure augments semantic relevance rather than overriding it.

\subsection{Memory packing and context assembly}
\label{sec:packing}

Retrieved nodes are bucketed by type with per-type caps (example - facts: 60, reflections: 20, episodes: 80) to prevent any specific type of node from dominating the context.
\new{This per-type budget keeps episodes, which have naturally high embedding similarity to conversational queries, from dominating the retrieval set and displacing facts.}
Finally the memory pack is trimmed proportionally within each bucket based on node scores to fit within an overall word budget \new{of 1{,}000 words}.
Episodes are sorted chronologically for temporal coherence.

\subsection{\rev{GRAFT: offline graph repair by answer memoization}}
\label{sec:graft}

\rev{Knowledge graph construction is a one-time process. If the extraction LLM misses a fact or omits a linking concept, the gap persists, and every later question that depends on it retrieves an incomplete context.
GRAFT closes such gaps between sessions rather than during answering.
It is self-supervised, since it never sees a reference answer and works only from the question, the memory retrieved for it, and its own hypothesis.
It also adds no LLM call to the answer path, because the answer prompt is untouched and repair runs asynchronously over the log of (question, retrieved memory) pairs.}

\paragraph{\rev{Adequacy triage.}}
\rev{One short classification call per logged question judges whether the retrieved memory, rather than the answer produced from it, contains what the question asks for, \new{and questions judged insufficient are queued for repair}.
\new{Judging the memory rather than the answer matters because answers are written to read confidently and can hide missing evidence.}}

\paragraph{\rev{Bounded ReAct investigation.}}
\rev{A queued question is handed to a ReAct agent~\citep{yao2023react} that alternates Thought, Action and Observation over a single tool, \textsc{search\_memory}, which runs GAAMA's ordinary retrieval (Section~\ref{sec:retrieval}) \new{with a smaller budget} on a sub-query the agent writes.
Each turn it must name the angle it is probing, namely entity, temporal, concept, relational, or episodic, and say why the previous observation fell short, which pushes successive probes to approach the question from complementary directions instead of rephrasing the same one.
\new{The loop is hard-bounded, with at most $I$ turns of one search each and a no-progress guard} that stops it once two consecutive searches return no new nodes.
If it reaches a bound without committing, a single further call forces a decision to answer or give up, based on the evidence already gathered.}

\paragraph{\rev{Grounded write-back.}}
\rev{An answer is memoized only if it clears three gates: a confidence floor of 0.6; an entailment check in which a verifier LLM must confirm that the answer follows from the specific nodes the agent cited; and rejection of hedged statements and of near-duplicates of \new{stored nodes (cosine similarity of at least 0.90)}.
What is written is one fact node, the answer restated as a standalone claim, with belief $b = \min(0.80, \text{confidence})$, a DERIVED\_FROM edge to each cited source episode and an ABOUT\_CONCEPT edge to each named concept.
No new node or edge type is introduced, so the memoized answer participates in ordinary retrieval and PPR traversal like any other fact, while its capped belief (Equation~\ref{eq:score}) ranks it below most extraction-time facts of equal relevance.
The graph therefore becomes a self-populating answer cache: the multi-angle investigation is paid for once, offline, and is recovered by every later question that touches the same material.}

\section{Experimental Setup}
\label{sec:experimental}

\subsection{Benchmarks}

\paragraph{LoCoMo-10.}
We evaluate on LoCoMo-10~\citep{maharana2024locomo}: 10 multi-session conversations with 1,540 questions across four categories.
Conversations vary in length (81--199 questions each), topic diversity, and structural complexity.

\paragraph{MemoryArena.}
\new{We also evaluate on \upd{two} MemoryArena datasets~\citep{memorybench2025}, described in Section~\ref{sec:memorybench}.}

\subsection{Evaluation protocol}
\label{sec:eval_protocol}

Our evaluation follows a \textbf{generate-then-judge} protocol:
(1)~retrieve relevant memories into a 1,000-word context budget;
(2)~generate an answer with \new{an} LLM based on query and retrieved memory;
(3)~score the generated answer with \new{an} LLM judge measuring key information coverage with respect to the reference answer.
The judge produces a fractional reward:
\begin{equation}
r = \frac{\text{number of reference key information found in hypothesis}}{\text{total key information in reference answer}}
\label{eq:reward}
\end{equation}
A score of 1.0 means all key information \new{is} present; partial scores reflect partial coverage.
The judge does \emph{not} penalize for extra information beyond the reference answer, avoiding unfair disadvantage to high-recall systems.

\subsection{Models and configuration}
\label{sec:config}

All experiments use GPT-4o-mini for extraction, generation, and judging (temperature=0), and text-embedding-3-small (1536-dim) for embeddings.
PPR configuration: $\alpha = 0.6$, max \rev{100} iterations, convergence tolerance $10^{-6}$, expansion depth $d = 2$,  $k = 40$ KNN seeds.
Scoring: $w_{\text{ppr}} = 0.1$, $w_{\text{sim}} = 1.0$.
\rev{Belief: facts and reflections carry the confidence assigned by the extraction LLM, while episodes and concepts have $b = 1$.
In the LoCoMo graph most facts have $b = 0.95$, within a range of 0.80 to 1.0, and reflections range from 0.5 to 0.9.
GRAFT facts are capped at $b = 0.80$, below most extraction-time facts, so they win a retrieval slot only when clearly more relevant.}
Budget: max\_facts=60, max\_reflections=20, max\_episodes=80, max\_memory\_words=1000.
\new{GRAFT runs at most $I = 5$ ReAct turns with one memory search per turn, and each search returns up to 5 facts and 10 episodes within 600 words.
An answer is written back only if its confidence is at least 0.6 and its cosine similarity to every stored node is below 0.90.}

\section{Results}
\label{sec:results}

\subsection{Main results on LoCoMo-10}
\label{sec:main_results}

Table~\ref{tab:main_results} reports mean reward across systems.
\new{GAAMA has the best overall score (79.1\%), 0.9~pp above HippoRAG~2, the strongest external system, and 4.2~pp above RAG, and offline GRAFT raises it to 79.4\%.}

\begin{table}[t]
\caption{Mean reward (\%) on LoCoMo-10 by question category. Best in \textbf{bold}. \rev{Overall is the question category weighted mean over all 1,540 questions. The GAAMA\,+\,GRAFT row is the same pipeline answering from the  graph after offline repair; the unrepaired pass of that run scores 79.1 overall, matching the GAAMA row. }\new{ For the post-repair score, answers that are word-for-word identical before and after repair receive the same score (Section~\ref{sec:graft_eval}).}}
\label{tab:main_results}
\centering
\resizebox{\textwidth}{!}{%
\begin{tabular}{@{}lccccc@{}}
\toprule
System & Multi-hop (Cat1) & Temporal (Cat2) & Open Domain (Cat3) & Single Hop (Cat4) & Overall \\
\midrule
A-Mem & \rev{48.0} & \rev{63.4} & \rev{36.6} & \rev{68.5} & \rev{61.7} \\
Nemori & 49.4 & 45.0 & 36.8 & 57.4 & 52.1 \\
Mem0 & \rev{72.8} & \rev{44.6} & \rev{43.4} & \rev{79.2} & \rev{68.6} \\
\rev{MAGMA} & \rev{51.9} & \rev{63.5} & \rev{41.7} & \rev{75.4} & \rev{66.5} \\
\rev{HyperMem} & \rev{62.2} & \rev{53.4} & \rev{47.0} & \rev{68.7} & \rev{63.0} \\
\rev{HippoRAG~2} & \rev{71.1} & \rev{72.7} & \rev{\textbf{\new{47.9}}} & \rev{86.0} & \rev{78.2} \\
RAG Baseline & 67.5 & 59.0 & 44.6 & \rev{86.9} & 74.9 \\
\midrule
GAAMA (ours) & 72.4 & \rev{\textbf{74.3}} & \rev{47.0} & \rev{86.8} & 79.1 \\
GAAMA + GRAFT & \rev{\textbf{73.0}} & \rev{73.3} & \rev{46.2} & \rev{\textbf{87.6}} & \textbf{79.4} \\
\bottomrule
\end{tabular}%
}
\end{table}

\paragraph{Analysis by category.}
\new{\textbf{Multi-hop (Cat1):} GAAMA, GRAFT, Mem0, and HippoRAG~2 form a leading group among which GRAFT scores the highest, and GAAMA leads RAG by 4.9~pp because its facts and reflections condense information that RAG must assemble from raw turns.} 

\new{\textbf{Temporal (Cat2):} GAAMA leads and improves over RAG by 15.3~pp, while GRAFT leaves temporal accuracy essentially unchanged.}
Conversation timestamps are supplied to all methods including RAG, so the improvement is not due to timestamp availability.
Rather, GAAMA's fact extraction resolves vague temporal references (e.g., ``last week'' $\to$ a specific date) and NEXT edge chains preserve episode ordering, enabling temporal sequence reasoning.

\new{\textbf{Open Domain (Cat3):} this category remains hard for every system, and GAAMA is within 1~pp of the best.}

\new{\textbf{Single Hop (Cat4):} GAAMA, GRAFT, RAG, and HippoRAG~2 perform comparably, since the answer usually sits in one message that embedding search finds directly.}

\subsection{GRAFT evaluation}
\label{sec:graft_eval}

\rev{Adequacy triage flags about one question in seven (224 of 1,540) as poorly served by retrieval, and these are the questions that actually fail. Flagged questions score about half as well as the rest (47\% versus 85\%), and although they are only 15\% of all questions, they produce 40\% of the answers that score zero.}

\rev{For the 224 flagged questions, the agent finds a supported answer for 82 and writes 40 new facts to the graph. The rest find nothing or repeat facts the graph already holds. About half of the written facts match the reference answer (22 of 40), \new{which is why GRAFT caps their belief so they are} retrieved only when clearly more relevant. Once written, they appear in the retrieved memory of 345 questions, more than one in five.}

\rev{Measuring the effect of 40 facts on 1,540 questions needs care, because re-running the same setup already moves the overall score by about 0.3~pp. The few temporal questions that dropped in the end-to-end run had unchanged retrieved memory, so the drop comes from judge and generation noise rather than from GRAFT, and we give word-for-word identical answers the same score on both sides. For a clean measurement we keep everything fixed except the written facts, score each side three times, and compare only the 470 questions whose retrieved memory changes (Table~\ref{tab:graft_attr}). Reward on these questions rises from 78.3\% to 79.5\%, and the gain sits on the 345 questions that retrieve a written fact ($+$1.6~pp), while questions that were only re-ranked do not change. The overall gain is small ($+$0.35~pp) because GRAFT wrote only 40 facts, but its value lies in what it does to the memory. It turns retrieval failures into facts the graph keeps, it adds no work when answering because repair runs offline, and each fact written once serves every later question that needs it.}

\begin{table}[t]
\caption{\rev{GRAFT effect with only its 40 written facts toggled, everything else fixed, scored on questions whose retrieved memory changes. Columns show the number of changed questions, those with a GRAFT fact ranked among retrieved items, mean reward (\%) on the changed questions without and with GRAFT, their difference, that difference spread over the whole category, and questions that improved/regressed. The Overall row pools all changed questions.}}
\label{tab:graft_attr}
\centering\small
\resizebox{\textwidth}{!}{%
\begin{tabular}{@{}lrrrrrrr@{}}
\toprule
Category & $n$ changed & Retrieve fact & w/o GRAFT & w/ GRAFT & $\Delta$ (changed) & $\Delta$ (full) & Impr./Regr. \\
\midrule
Cat1 (Multi-hop) & 86 & 65 & 72.0 & 73.5 & $+$1.5 & $+$0.45 & 7\,/\,3 \\
Cat2 (Temporal) & 98 & 69 & 75.8 & 75.5 & $-$0.3 & $-$0.09 & 8\,/\,7 \\
Cat3 (Open Domain) & 31 & 26 & 41.2 & 39.3 & $-$2.0 & $-$0.64 & 1\,/\,3 \\
Cat4 (Single Hop) & 255 & 185 & 86.0 & 87.9 & $+$2.0 & $+$0.59 & 17\,/\,6 \\
\midrule
\textbf{Overall} & \textbf{470} & \textbf{345} & \textbf{78.3} & \textbf{79.5} & $+$\textbf{1.1} & $+$\textbf{0.35} & \textbf{33\,/\,19} \\
\bottomrule
\end{tabular}%
}
\end{table}

\rev{Single-hop and multi-hop questions carry the gain ($+$0.59 and $+$0.45~pp on the full set).
Temporal questions are flat, and open-domain questions lose 0.64~pp over 31 affected questions---too small a set to separate from noise, but consistent with the write-precision figure above: an open-domain question is the hardest case for a single memoized statement to serve, since what it asks for is rarely one standalone claim.}

\paragraph{\rev{Repair offline, or reason at query time?}}
\rev{The same agent could instead answer flagged questions directly at query time. Both options give about the same accuracy, so the choice comes down to cost and reuse. Answering at query time adds a search of up to five LLM steps for one question in seven, which makes those answers slower and more expensive, and the result helps only that one question. Offline repair keeps every answer to one retrieval and one LLM call, and each fact it writes is reused by later questions that need it. GAAMA therefore repairs offline.}

\subsection{\new{Ablation study}}
\label{sec:ablation}

\new{To isolate the contribution of graph structure and of the concept layer, we compare three configurations (Table~\ref{tab:ablation-cat}).
\textbf{Semantic} ranks GAAMA's memory by cosine similarity alone, with no graph traversal.
\textbf{Entity graph} replaces GAAMA's facts, concepts, and reflections with entity-relation triples and keeps the same retrieval.
\textbf{GAAMA (PPR=0.1)} is the full system with graph-augmented retrieval ($w_{\text{ppr}}{=}0.1$, $w_{\text{sim}}{=}1$).}

\rev{\new{Semantic and GAAMA} use identical memory stores, budgets, and generation. They differ in the ranking function and in one scoring detail that we state explicitly: the Semantic arm ranks candidates by cosine similarity alone, whereas \new{GAAMA} additionally scales each candidate by its node belief score.}

The semantic baseline here already benefits from GAAMA's three-step LTM construction (facts, reflections, episodes); it is \emph{not} the flat RAG baseline.
This ablation isolates specifically whether \new{belief-weighted} PPR provides additional value over semantic search on the same constructed LTM.

\new{Compared with Semantic, GAAMA (PPR=0.1)} improves overall reward by \rev{+1.3~pp (77.8\% $\to$ 79.1\%)}, with the largest gain on temporal questions (\rev{+3.2~pp}).
\new{The overall gain is statistically significant, at $+$1.2~pp on the 1{,}526 questions both runs answer (paired bootstrap 95\% CI 0.2 to 2.3), with GAAMA ahead in 9 of 10 conversations (paired $t(9){=}3.94$).}
\rev{Table~\ref{tab:ablation-cat} reports the breakdown by question category.}

\begin{table}[t]
\caption{\new{Ablation by question category (mean reward, \%). Semantic ranks GAAMA's memory by similarity alone, without the graph. Entity graph replaces GAAMA's facts, concepts, and reflections with entity-relation triples over the same episodes, retriever, budget, and judge. GAAMA (PPR=0.1) is the full run reported in Table~\ref{tab:main_results}.}}
\label{tab:ablation-cat}
\centering
\small
\resizebox{\textwidth}{!}{%
\begin{tabular}{@{}lccccc@{}}
\toprule
\new{Setting} & Multi-hop (Cat1) & Temporal (Cat2) & Open Domain (Cat3) & Single Hop (Cat4) & Overall \\
\midrule
Semantic & 73.9 & 71.1 & 45.8 & 85.4 & 77.8 \\
\new{Entity graph} & \new{73.2} & \new{61.7} & \new{45.7} & \new{86.0} & \new{76.1} \\
\midrule
\new{GAAMA (PPR=0.1)} & 72.4 & \rev{74.3} & \rev{47.0} & \rev{86.8} & 79.1 \\
\bottomrule
\end{tabular}%
}
\end{table}

\rev{\textbf{Cat2 (Temporal) benefits most from PPR} (+3.2~pp), with 22 questions improving and 12 regressing.}
PPR seeds spread probability through all five edge types simultaneously---a temporal query benefits from NEXT edges discovering adjacent episodes, concept edges surfacing date-bearing facts from the same topic, and provenance edges linking back to source episodes with timestamps.

\rev{\textbf{Cat1 (Multi-hop) is the only net-negative category} ($-$1.5~pp), but the change is not statistically significant, with 35 questions regressing and 31 improving.}

\textbf{Cat3 (Open Domain)} gains \rev{+1.2~pp (10 improved, 5 regressed)}, but the small category size (96 questions) limits statistical confidence.

\new{\textbf{Concept vs.\ entity graph.} Replacing facts, concepts, and reflections with entity-relation triples over the same episodes, retriever, budget, and judge lowers reward by 3.0~pp (79.1\% to 76.1\%; paired bootstrap 95\% CI 1.4 to 4.6; paired $t(9){=}4.42$, $p{=}0.002$, with GAAMA ahead in 9 of 10 conversations; Table~\ref{tab:ablation-cat}).
Most of the loss comes from content.
Splitting a message into triples drops context that a fact keeps, and the clearest measured case is time, where only 3.4\% of the triples in context carry a date against 52.4\% of GAAMA's facts, giving a 12.6~pp drop on temporal questions while the other categories tie.
The hubs act differently.
The two speakers link to up to 336 nodes per conversation, while no concept exceeds 28, which makes propagation uninformative .}

\subsection{MemoryArena evaluation}
\label{sec:memorybench}

\upd{We also evaluate on MemoryArena~\citep{memorybench2025}, where questions arrive in sequences and later steps depend on earlier answers.
Its tasks were built for agents that use tools, so we keep only what memory alone can answer.
We drop Web Shopping, where every step selects a product by a price or rating that the prompt never shows.
\emph{Group Travel Planner} (270 sequences, 1{,}869 questions) adds travelers whose plans must respect earlier itineraries.
Most itinerary items can be copied from the plan in the prompt (73\%) or need a database search (18\%), so we score only the places a traveler asks to share with an earlier traveler, which only memory can supply.
This leaves 1{,}550 itinerary items across 960 of the 1{,}599 questions after the cold-start step.
\emph{Progressive Search} (221 sequences, 1{,}641 questions) adds one clue per step toward a target entity and is kept whole, since from the second step every answer is in memory.}

\paragraph{Evaluation protocol.}
\upd{At each step the system retrieves from its memory using the current question, answers it, and is scored by the GPT-4o-mini judge, after which the reference answer is added to memory.
For each dataset category, an independent memory is created, ensuring no information leaks across datasets.
The cold-start step Q0 is excluded.
All memory systems receive at most 1{,}000 words of retrieved memory.
Before this cap, GAAMA's budget is 40 facts, 8 reflections, and 240 episodes on Progressive Search, and 240 episodes on Travel Planner.
On Travel Planner, the judge checks whether each memory-determined venue appears in the generated itinerary, and all systems are compared on the 960 questions they share.}

\begin{table}[t]
\caption{\upd{MemoryArena results. Progressive Search reports mean reward (\%) over all steps after the cold-start step Q0. Travel Planner reports the share (\%) of memory-determined venues found in the generated itinerary, over 1{,}535 items in the 960 questions shared by all systems. Avg.\ is the mean of the two columns. Best in bold.}}
\label{tab:memorybench_merged}
\centering\small
\begin{tabular}{@{}lccc@{}}
\toprule
System & \upd{Prog.\ Search} & \upd{Travel Planner (memory items)} & \upd{Avg.} \\
\midrule
HippoRAG~2 & \upd{76.2} & \upd{46.2} & \upd{61.2} \\
Mem0 & \upd{69.9} & \upd{17.9} & \upd{43.9} \\
Nemori & \upd{74.0} & \upd{12.8} & \upd{43.4} \\
A-Mem & \upd{76.3} & \upd{40.9} & \upd{58.6} \\
\midrule
GAAMA (ours) & \upd{\textbf{76.6}} & \upd{\textbf{48.6}} & \upd{\textbf{62.6}} \\
\bottomrule
\end{tabular}
\end{table}

\paragraph{Results.}
\upd{On Progressive Search, GAAMA has the best score, ahead of Nemori by 2.6~pp (95\% CI 1.3 to 3.9) and Mem0 by 6.7~pp (CI 5.3 to 8.3), and statistically tied with A-Mem and HippoRAG~2 (Table~\ref{tab:memorybench_merged}).
On Travel Planner, GAAMA recovers 48.6\% of the memory-determined venues, ahead of A-Mem by 7.6~pp (CI 4.9 to 10.3), far ahead of Mem0 and Nemori, and slightly above HippoRAG~2.
Mem0 and Nemori fall far behind because the needed venue reaches their prompt for only 15--17\% of items, as Nemori's narrative episodes rarely keep venue names and Mem0 stores summaries of the options rather than the choice.}

\section{Conclusion}
\label{sec:conclusion}

\new{We presented GAAMA, a graph-augmented associative memory that stores conversations as episodes, self-contained facts, and reflections, links them through topic concepts instead of entity nodes, and retrieves with a hybrid of semantic search and Personalized PageRank.
On LoCoMo-10, GAAMA reaches 79.1\%, the best average among the compared methods, and offline GRAFT repair raises it to 79.4\% without adding work at query time.
Most of the gain comes from the memory representation, with graph-augmented retrieval adding a further gain, and replacing GAAMA's facts with entity-relation triples lowers reward by 3.0~pp because the triples lose context that facts keep.}

\new{Several directions remain open.
Retrieval and edge-type weights are set by hand and shared by all queries, and learning them from retrieval feedback or adapting them to the question type could raise the graph's contribution.
Because every fact links to the episode it came from, the context builder could choose, for each piece of evidence, the level that best serves the question within the budget, sending the compact fact when it suffices and the full episode when exact details are needed, rather than sending both.
GRAFT writes facts that match the reference only about half the time, so stronger verification before write-back would make repair more reliable.
Finally, consolidating contradictory or outdated facts and updating the graph incrementally are needed to scale GAAMA to hundreds of sessions.}


\bibliographystyle{plainnat}

\end{document}